\documentclass[conference]{IEEEtran}
\IEEEoverridecommandlockouts

\usepackage{cite}
\usepackage{amsmath,amssymb,amsfonts}
\usepackage{algorithmic}
\usepackage{graphicx}
\usepackage{subfigure}
\usepackage{textcomp}
\usepackage{xcolor}
\usepackage{algorithm}
\usepackage{amssymb}
\usepackage{tabularx}
\usepackage{multirow}
\usepackage{array}
\usepackage{booktabs}

\usepackage{url}
\def\BibTeX{{\rm B\kern-.05em{\sc i\kern-.025em b}\kern-.08em
    T\kern-.1667em\lower.7ex\hbox{E}\kern-.125emX}}

\makeatletter
\renewcommand{\@makecaption}[2]{%
  \vskip\abovecaptionskip
  \sbox\@tempboxa{\normalsize\textbf{#1}. \normalsize#2}%
  \ifdim \wd\@tempboxa >\hsize
    \normalsize\textbf{#1}. \normalsize#2\par
  \else
    \global \@minipagefalse
    \hb@xt@\hsize{\hfil\box\@tempboxa\hfil}%
  \fi
  \vskip-0.5cm}  
\makeatother

\begin{document}

\title{Adaptive Spatiotemporal Augmentation for Improving Dynamic Graph Learning\\
}

\author{
\IEEEauthorblockN{Xu Chu\textsuperscript{1,†}, Hanlin Xue\textsuperscript{1,†}, Bingce Wang\textsuperscript{1}, Xiaoyang Liu\textsuperscript{2}, Weiping Li\textsuperscript{1,*}, Tong Mo\textsuperscript{1}, Tuoyu Feng\textsuperscript{1}, Zhijie Tan\textsuperscript{1}}
\IEEEauthorblockA{\textsuperscript{1}\textit{School of Software and Microelectronics, Peking University}, Beijing, China \\
\textsuperscript{2}\textit{EEIS Department, University of Science and Technology of China}, Anhui, China \\
Emails: chuxu@stu.pku.edu.cn, colinneverland@stu.pku.edu.cn, wangbingce@stu.pku.edu.cn, 
liuxiaoyang@mail.ustc.edu.cn,\\
wpli@ss.pku.edu.cn, motong@ss.pku.edu.cn, fengty\_dky1@163.com, besttangent@stu.pku.edu.cn}
\thanks{\textsuperscript{†}These authors contributed equally to this work.}
\thanks{\textsuperscript{*}Corresponding author.}
}

\maketitle

\begin{abstract}
Dynamic graph augmentation is used to improve the performance of dynamic GNNs. Most methods assume temporal locality, meaning that recent edges are more influential than earlier edges. However, for temporal changes in edges caused by random noise, overemphasizing recent edges while neglecting earlier ones may lead to the model capturing noise. To address this issue, we propose STAA (SpatioTemporal Activity-Aware Random Walk Diffusion). STAA identifies nodes likely to have noisy edges in spatiotemporal dimensions. Spatially, it analyzes critical topological positions through graph wavelet coefficients. Temporally, it analyzes edge evolution through graph wavelet coefficient change rates. Then, random walks are used to reduce the weights of noisy edges, deriving a diffusion matrix containing spatiotemporal information as an augmented adjacency matrix for dynamic GNN learning. Experiments on multiple datasets show that STAA outperforms other dynamic graph augmentation methods in node classification and link prediction tasks.
\end{abstract}


\section{Introduction}
\begin{figure}[t]
\centering
\includegraphics[width=0.45\textwidth]{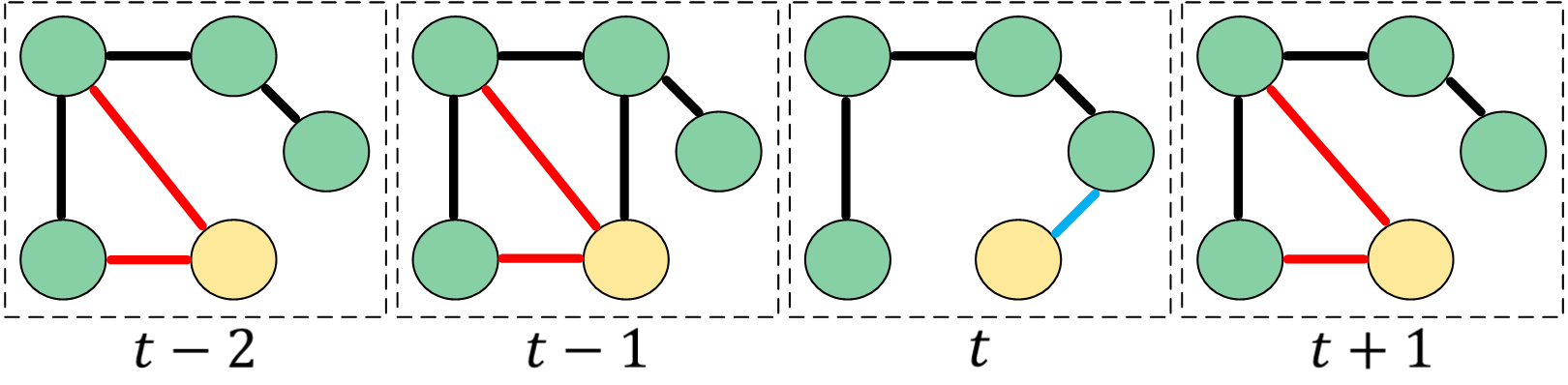} 
\caption{Example of overemphasizing the recent edges.}
\label{intro}
\end{figure}

\footnotetext{Our code is available at https://github.com/ColinNeverLand/STAA}
To enhance the generalization ability and performance of Graph Neural Networks (GNNs), researchers use graph augmentation to achieve better graph learning~\cite{Zhao2020DataAF,Zhu2020GraphCL,Zhang2022SpectralFA,Sui2022UnleashingTP}. Recent studies combine GNNs with recurrent networks and transformers, producing dynamic GNNs~\cite{Seo2016StructuredSM,Pareja2019EvolveGCNEG,Sankar2020DySATDN,Zhu2023WinGNNDG,liu2024todynet}. Extending augmentation to dynamic graphs has yielded significant results~\cite{Wang2021AdaptiveDA,lee2023time,Chen2023TemporalGR,tian2024latent}. However, dynamic graphs contain dynamic noise, which is characterized by temporally evolving structures
and uncertain changes~\cite{Zhang2023RDGSLDG}. Most augmentation methods rely on temporal locality assumption, which posits that recent edges are more important than earlier ones for augmenting node representations~\cite{Chen2023TemporalGR}. This assumption does not apply to temporal changes in edges (i.e., structure) caused by noise. As shown in Figure~\ref{intro}, the link prediction task predicts edges on the graph at time $t+1$ based on graph snapshots at time $t$ and earlier. Augmentation methods that excessively focus on recent edges (edges of yellow nodes at time $t$) while neglecting earlier edges (edges of yellow nodes at times $t-1$ and $t-2$) may lead the model to capture noise (blue edge) rather than effective spatiotemporal information (red edges). To identify and suppress noisy edges, we approach from the node perspective, categorizing the nodes in the dynamic graph into active nodes (frequently changing structure, such as the yellow nodes in Figure~\ref{intro}) and inert nodes (stable structure). Active nodes frequently alter their interactions with other nodes, making them more susceptible to noisy edges~\cite{Zhang2023RDGSLDG}.

In this paper, we propose STAA (\underline{S}patio\underline{T}emporal \underline{A}ctivity-\underline{A}ware Random Walk Diffusion), a novel discrete-time dynamic graph augmentation method. STAA assesses nodes' spatiotemporal characteristics, calculating activity coefficients. Nodes with high coefficients are active nodes. Specifically, in spatial domain, STAA analyzes critical topological positions through graph wavelet coefficients, while in temporal domain, it analyzes edge evolution through graph wavelet coefficient change rates. Subsequently, the random walk defined on the dynamic graph suppresses the active nodes' preference for recent edges and increases the temporal walking probability of earlier edges to obtain a larger temporal receptive field, thereby generating a diffusion matrix that reduces the weight of noise edges. We use this diffusion matrix as an augmented adjacency matrix for the dynamic graph. Our main contributions are: (a) We introduced a method for evaluating the spatiotemporal activity of nodes based on wavelet coefficients and their rate of change within a time window. (b) We propose STAA, a model-agnostic dynamic graph augmentation method. STAA suppresses noise and enhances the spatiotemporal information of dynamic graphs. (c) Extensive experiments on multiple popular dynamic graph datasets demonstrate that, compared to other graph augmentation methods, using STAA to enhance dynamic graphs enables GNNs to achieve superior performance in node classification and link prediction tasks.

\begin{figure}[t]
    \centering
    \subfigure[Low and high-frequency coefficients on the same graph.]{
        \includegraphics[width=0.45\textwidth]{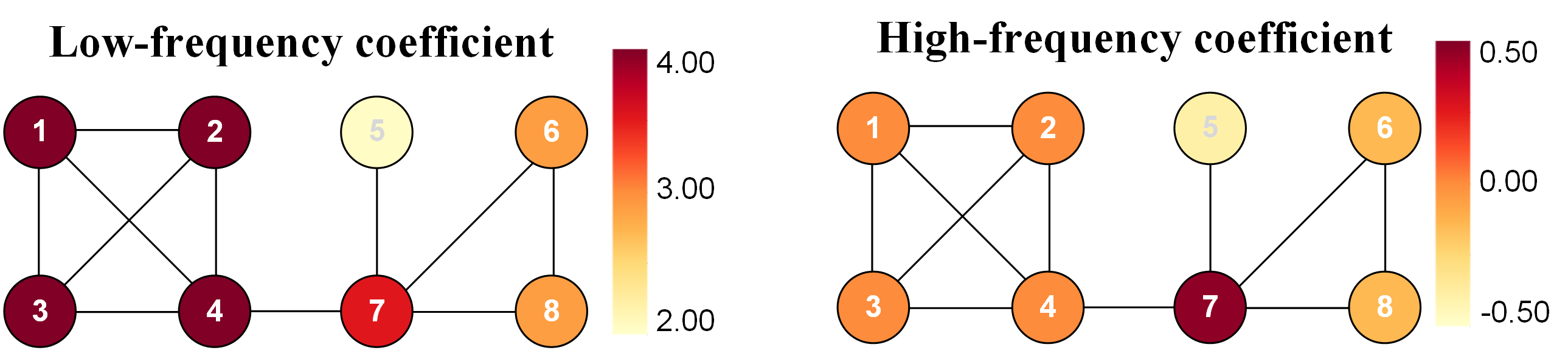}
        \label{fig:subfig1}
    }
    \subfigure[Changes in low-frequency coefficients with graph structure changes.]{
        \includegraphics[width=0.45\textwidth]{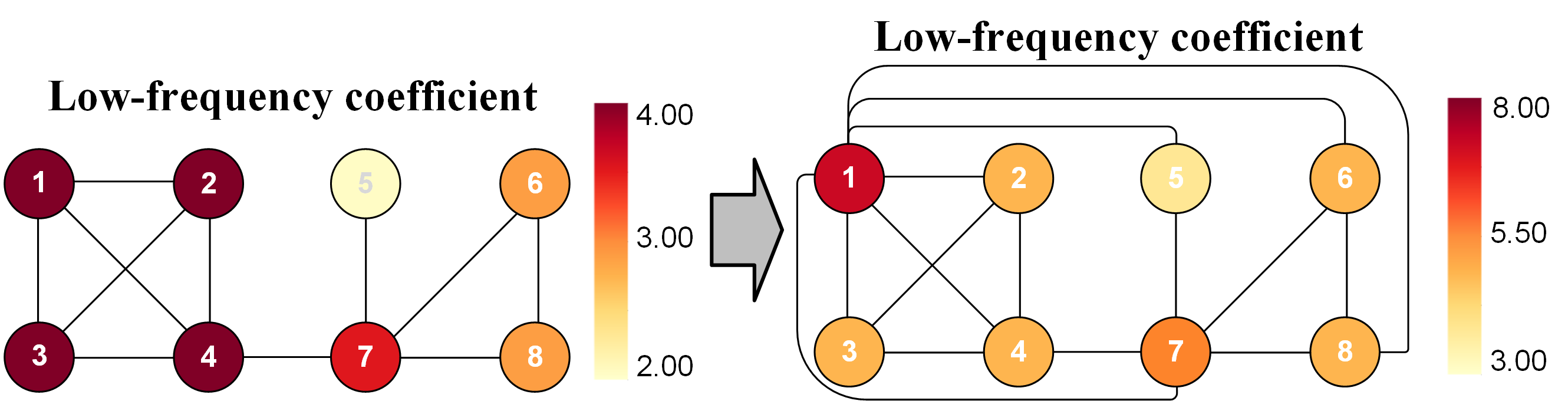}
        \label{fig:subfig2}
    }
    \caption{Heatmap of wavelet coefficients.}
    \label{gwt_intro}
\end{figure}

\section{Preliminaries}
In this paper, we concentrate on the discrete-time dynamic graph~\cite{Kazemi2019RepresentationLF}, defined as a series of snapshots $\mathcal{G}=\left\{\mathcal{G}_1, \mathcal{G}_2, \ldots, \mathcal{G}_T\right\}$, where $T$ represents the total number of snapshots. The snapshot at time $t$, denoted as $\mathcal{G}_t = \left(\mathcal{V}_t, \mathcal{E}_t,\mathbf{F}_t\right)$, is a graph with a shared set $\mathcal{V}$ of nodes and a set $\mathcal{E}_t$ of edges, where $n=|\mathcal{V}|$ is the number of nodes. $\mathbf{F}_t \in \mathbb{R}^{n \times m}$ is node feature matrix where $m$ is the feature dimension. The adjacency matrix corresponding to the edge set $\mathcal{E}_t$ is denoted by $\mathbf{A}_t$. 

\textbf{Graph Wavelet Transform (GWT).} 
For GWT~\cite{Hammond2009WaveletsOG}, the graph Laplacian is given by $\mathbf{L}=\mathbf{D}-\mathbf{A}$, where $\mathbf{D}$ is the degree matrix. $L$ is real symmetric and positive semi-definite, with a rank less than $n$. Consequently, it possesses a complete set of orthogonal eigenvectors $\left\{u_i\right\}, i \in\{1, \ldots, n\}$, corresponding to non-negative real eigenvalues $\left\{\lambda_i\right\}$, $i \in\{1, \ldots, n\}$. 
The wavelet coefficient of node $v_j$ for function $f$ at scale $\omega$ is defined as follows:

\begin{equation}\label{eq1}
\begin{aligned}
    W_f(\omega,j) &= \sum_{i=1}^n g(\omega\lambda_i) \hat{f}(i)u_i(j), \\
    \hat{f}(i) &= u_i^\top \cdot f = \sum_{j=1}^n u_i(j)f(j),
\end{aligned}
\end{equation}
where $u_i^\top$ is the transpose of $u_i$, $u_i(j)$ is the $j$-th element of $u_i$, $f(j)$ is the value of $f$ at node $v_j$, $f$ is a real-valued function defined on the nodes of $\mathcal{G}_t$. The scales used to generate the wavelet kernels $g(\omega x)$ are $\omega_0, \omega_1, \ldots, \omega_{r-1}$, where $r$ is the number of scales, $\omega_0$ and $\omega_{r-1}$ correspond to the highest and lowest frequency Fourier modes, respectively. $g$ is the filter kernel on $\mathbb{R}^+$. We use the same kernel function as in~\cite{Col2018WaveletBasedVA}.


\textbf{Analysis of Node Dynamics.} As shown in Figure~\ref{gwt_intro}, we compute graph wavelet coefficients using node degrees as the function $f$ (i.e., $f(j) = d_j$, where $d_j$ represents the degree of node $v_j$), and $r$ is 6. Figure~\ref{gwt_intro}(a) compares the low-frequency (corresponding to scale $\omega_5$) and high-frequency (corresponding to scale $\omega_0$) coefficients. Nodes \textit{1-4}, with larger low-frequency content, have more consistent neighborhood degrees, while node \textit{7}, with larger high-frequency content, shows larger differences in degree distribution within the neighborhood. Figure~\ref{gwt_intro}(b) illustrates how the low-frequency coefficients change when the graph structure is altered. The change in the degree of node \textit{1} affects its low-frequency coefficients and those of its one-hop neighbors.

Based on the above analysis, we propose \textbf{Observation}: (a) On a single graph snapshot, low-frequency coefficients correspond to smooth graph signal patterns. Nodes with higher high-frequency coefficients exhibit unsmooth neighborhood signal distributions, occupy critical topological positions, are susceptible to noise~\cite{Sandryhaila2013DiscreteSP}, and are likely to be spatially active. (b) Across different snapshots, nodes with frequently changing signals show higher low-frequency coefficient change rates and are more likely to be temporally active. 

\begin{figure}[t]
\centering
\includegraphics[width=0.48\textwidth]{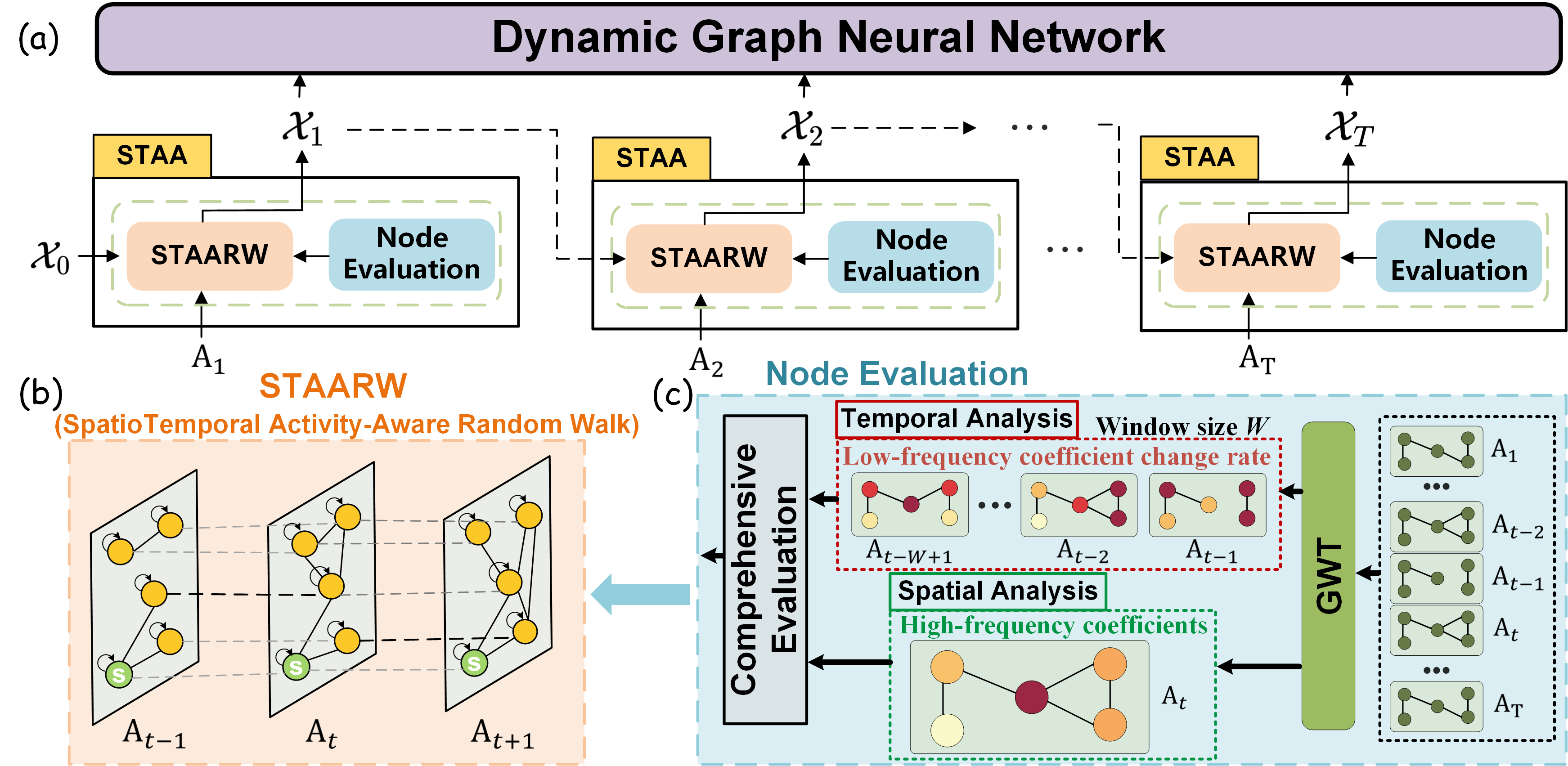} 
\caption{Overview of the proposed STAA.}
\label{STAA}
\end{figure}

Note that we choose node degree as our function $f$ (i.e., graph signal) to characterize neighborhoods, as it is a fundamental measure of connectivity that directly reflects the graph's structural dynamics.
\begin{table*}[t]
\renewcommand{\arraystretch}{0.9}

\resizebox{\textwidth}{!}{
\centering
\begin{tabular}{@{}*{10}{c}@{}}
\toprule[1.5pt]
\multirow{2}{*}{\textbf{AUC}} & \multicolumn{3}{c}{\textbf{BitcoinAlpha}}           & \multicolumn{3}{c}{\textbf{WikiElec}}               & \multicolumn{3}{c}{\textbf{RedditBody}}             \\
\cmidrule(lr){2-4} \cmidrule(lr){5-7} \cmidrule(lr){8-10}
& \textbf{GCN} & \textbf{GCRN} & \textbf{EvolveGCN} & \textbf{GCN} & \textbf{GCRN} & \textbf{EvolveGCN} & \textbf{GCN} & \textbf{GCRN} & \textbf{EvolveGCN}\\ \midrule
NONE                 & 57.0$\pm$1.2 & 87.4$\pm$9.0 & 59.3$\pm$0.3 & 59.6$\pm$1.1 & 73.6$\pm$3.7 & 64.8$\pm$4.0 & 76.7$\pm$0.1 & 88.5$\pm$0.4 & 76.3$\pm$0.2 \\
\midrule
DropEdge*             & 56.3$\pm$1.0 & 73.9$\pm$2.2 & 57.4$\pm$0.9 & 50.1$\pm$1.0 & 56.0$\pm$9.3 & 47.9$\pm$6.4 & 73.0$\pm$0.4 & 77.0$\pm$1.7 & 71.9$\pm$0.7 \\
GDC*                  & 57.5$\pm$1.6 & 77.3$\pm$6.5 & 57.4$\pm$1.2 & 62.8$\pm$0.8 & 67.9$\pm$1.0 & 63.1$\pm$0.7 & 74.6$\pm$0.0 & 86.4$\pm$0.3 & 73.8$\pm$0.3 \\
MERGE                & 67.4$\pm$2.2 & 92.5$\pm$1.0 & 61.2$\pm$1.9 & 61.6$\pm$1.0 & 68.4$\pm$4.8 & 65.7$\pm$0.8 & 69.8$\pm$0.1 & 89.7$\pm$0.8 & 79.1$\pm$0.7\\
TIARA                & 76.7$\pm$1.5 & 94.2$\pm$0.7 & 77.2$\pm$1.3 & 68.5$\pm$0.4 & 70.8$\pm$5.1 & 68.9$\pm$0.4 & 80.2$\pm$1.0 & 89.8$\pm$0.8 & 81.8$\pm$0.3 \\
TGAC                & 68.1$\pm$1.3 & 78.2$\pm$7.7 & 69.9$\pm$1.6 & 63.5$\pm$1.3 & 75.1$\pm$6.2 & 63.0$\pm$0.2 & 76.1$\pm$0.8 & 81.2$\pm$2.8 & 77.0$\pm$0.0 \\
\midrule
\textbf{STAA}                 & \textbf{79.6$\pm$0.8} & \textbf{96.0$\pm$1.0} & \textbf{85.1$\pm$0.9} & \textbf{68.7$\pm$0.4} & \textbf{76.0$\pm$1.8} & \textbf{69.4$\pm$0.7} & \textbf{82.2$\pm$0.3} & \textbf{90.1$\pm$0.5} & \textbf{83.2$\pm$0.3} \\ \bottomrule[1.5pt]
\end{tabular}
}
\caption{Link prediction accuracy (AUC).}
\label{tab:link_pre}
\end{table*}
\section{METHOD}
The framework of STAA is illustrated in Figure~\ref{STAA}. Initially, for a dynamic graph $\mathcal{G}$ with adjacency matrices $\{\mathbf{A}_1, \ldots, \mathbf{A}_T\}$, STAA starts by processing the first adjacency matrix $\mathbf{A}_1$ and an initial diffusion matrix $\mathcal{X}_0$ as shown in Figure~\ref{STAA}(a). By utilizing node evaluations derived from graph wavelet transforms, STAA directs the SpatioTemporal Activity-Aware Random Walk (STAARW) to produce a diffusion matrix $\mathcal{X}_1$. At each subsequent time step $t$, STAA updates its input by incorporating the new adjacency matrix $\mathbf{A}_t$ and the last diffusion matrix $\mathcal{X}_{t-1}$ to generate the next diffusion matrix $\mathcal{X}_t$. To improve dynamic graph learning, the diffusion matrix $\mathcal{X}_t \in \mathbb{R}^{n \times n}$ at each time step $t$ supersedes $\mathbf{A}_t$ in training dynamic graph neural networks.
\subsection{Node Evaluation}\label{Node Evaluation}
In this section, Graph Wavelet Transform (GWT) is used to evaluate the spatiotemporal activity of nodes, i.e., to assess whether nodes are active or inert. For node $v_j$ on $\mathcal{G}_t$ and $r \in \mathbb{Z}^+, r \geqslant 2$, we define the low-frequency coefficient ${a}_{t,j}$ and high-frequency coefficient ${b}_{t,j}$ as:
\begin{equation}\label{eq3}
    {a}_{t,j} =\sum ^{r-1}_{l=\lfloor \frac{r}{2} \rfloor}e^{\lambda (l-r+1)}{W_f}(\omega_l,j),\
{b}_{t,j} =  \sum ^{\lfloor \frac{r}{2}\rfloor-1 }_{l=0}e^{-\lambda l}{W_f}(\omega_l,j),
\end{equation}
where $\lambda$ controls exponential decay magnitude, ${W_f}(\omega_l,j)$ is graph wavelet coefficient of node $v_j$ at time $t$ for scale $\omega_l$.

According to the \textbf{Observation}, nodes with large high-frequency coefficients on $\mathcal{G}_t$ have unsmooth signal distributions in their neighborhoods, indicating that at time $t$, these nodes are in topologically critical areas within their neighborhoods and are susceptible to neighborhood noise~\cite{Sandryhaila2013DiscreteSP}. This reflects \textbf{spatial} activity. Additionally, the higher the rate of change in the node's low-frequency coefficients between snapshots, the more \textbf{temporally} active the node is. Therefore, node activity is related to both time (low-frequency coefficient change rate) and space (high-frequency coefficients). The low-frequency coefficient change rate is defined as:
\begin{equation}\label{eq4}
\begin{aligned}
    \Delta a_{t,j} &= \frac{1}{W-1} \sum_{i=1}^{W-1} |a_{t-i+1,j} - a_{t-i,j}|, \\
    \hat{\Delta a_{t,j}} &= \frac{\Delta a_{t,j} - \mu(\Delta a_t)}{\sigma(\Delta a_t) + \epsilon},
\end{aligned}
\end{equation}
where $1\leq W\leq T$ is the size of the time window, $\mu(\Delta a_t)$ and $\sigma(\Delta a_t)$ are the mean and standard deviation of $\Delta a_{t,j}$ for all nodes on $\mathcal{G}_t$, respectively, and $\epsilon$ is a small constant to prevent division by zero. $\hat{\Delta a_{t,j}}$ is the normalized low-frequency coefficient change rate. The normalized high-frequency coefficient is defined as:
\begin{equation}\label{eq5}
    \hat{b_{t,j}} = \frac{b_{t,j} - \mu(b_t)}{\sigma(b_t) + \epsilon},
\end{equation}
where $\mu(b_t)$ and $\sigma(b_t)$ are the mean and standard deviation of $b_{t}$ for all nodes on $\mathcal{G}_t$, respectively. The spatiotemporal activity coefficient $\beta_{t,j}$ of node $v_j$ at time $t$ is defined as:
\begin{equation}\label{eq6}
\begin{aligned}
    \beta_{t,j} &= \delta \cdot \sigma(\hat{\tau_{t,j}}), \\
    \hat{\tau_{t,j}} &= \frac{\tau_{t,j} - \mu(\tau_t)}{\sigma(\tau_t) + \epsilon},
    \tau_{t,j} = \gamma \hat{\Delta a_{t,j}} + (1 - \gamma) \hat{b_{t,j}},
\end{aligned}
\end{equation}
where $\sigma(\cdot)$ represents the sigmoid function, $\delta$ is a scaling factor, and $\gamma$ is a gating factor used to balance the influence of the low-frequency coefficient change rate and high-frequency coefficient. $\mu(\tau_t)$ and $\sigma(\tau_t)$ are the mean and standard deviation of $\tau_{t}$ for all nodes on $\mathcal{G}_t$, respectively.

\begin{table*}[t]
\setlength{\tabcolsep}{3pt}

\resizebox{\textwidth}{!}{%
\begin{tabular}{@{}*{13}{c}@{}}
\toprule[1.5pt]
\multirow{2}{*}{\textbf{Macro F1}} & \multicolumn{3}{c}{\textbf{Brain}} & \multicolumn{3}{c}{\textbf{Reddit}} & \multicolumn{3}{c}{\textbf{DBLP-3}} & \multicolumn{3}{c}{\textbf{DBLP-5}} \\
\cmidrule(lr){2-4} \cmidrule(lr){5-7} \cmidrule(lr){8-10} \cmidrule(lr){11-13}
 & \textbf{GCN} & \textbf{GCRN} & \textbf{EvolveGCN} & \textbf{GCN} & \textbf{GCRN} & \textbf{EvolveGCN} & \textbf{GCN} & \textbf{GCRN} & \textbf{EvolveGCN} & \textbf{GCN} & \textbf{GCRN} & \textbf{EvolveGCN} \\
\midrule
NONE                               & 45.6$\pm$2.8          & 63.3$\pm$1.5          & 45.4$\pm$1.1          & 21.4$\pm$0.7          & 40.8$\pm$1.0          & 23.1$\pm$0.7          & 54.0$\pm$1.8          & 82.2$\pm$0.6          & 54.5$\pm$0.9          & 69.3$\pm$0.5          & 74.4$\pm$0.4          & 68.4$\pm$0.4          \\
\midrule
DropEdge*                           & 35.2$\pm$1.7          & 67.8$\pm$0.6          & 39.7$\pm$1.8          & 19.4$\pm$0.8          & 40.3$\pm$1.4          & 18.0$\pm$2.7          & 55.8$\pm$1.9          & 84.3$\pm$0.6          & 52.4$\pm$1.7          & 70.5$\pm$0.5          & 75.6$\pm$0.7          & 68.0$\pm$0.7          \\
GDC*                                & 63.2$\pm$1.2          & 88.0$\pm$1.5          & 67.3$\pm$1.3          & 17.5$\pm$2.3          & 41.0$\pm$1.6          & 18.5$\pm$2.8          & 53.4$\pm$2.1          & 84.7$\pm$0.5          & 52.8$\pm$2.2          & 70.0$\pm$0.7          & 75.5$\pm$1.2          & 69.1$\pm$1.0          \\
MERGE                              & 43.1$\pm$5.8          & 60.4$\pm$4.7          & 54.2$\pm$3.9          & 22.3$\pm$0.5          & 41.1$\pm$1.9          & 24.6$\pm$0.4          & 56.1$\pm$1.6          & 82.7$\pm$1.0          & 56.6$\pm$0.5          & 70.0$\pm$0.3          & 74.1$\pm$1.2          & 69.1$\pm$0.4          \\
TIARA                              & 68.2$\pm$0.6          & 90.3$\pm$2.2          & {72.1$\pm$0.4}          & 21.2$\pm$4.3          & 41.6$\pm$1.3          & 22.5$\pm$1.6          & 56.4$\pm$1.1          & 84.9$\pm$1.2          & 56.1$\pm$0.6          & 70.0$\pm$0.6          & 75.2$\pm$1.3          & 69.6$\pm$0.3          \\
TGAC                              & 42.3$\pm$1.5          & 27.2$\pm$2.6          & {36.5$\pm$1.1}          & 24.2$\pm$2.1          & 39.2$\pm$1.3          & 16.9$\pm$2.1          & \textbf{63.8$\pm$1.0}          & 82.0$\pm$0.8          & 53.3$\pm$0.3          & 70.0$\pm$0.7          & 75.0$\pm$0.8          & 67.9$\pm$0.6          \\
\midrule
\textbf{STAA}                      & \textbf{70.4$\pm$0.3} & \textbf{90.6$\pm$1.7} & \textbf{73.0$\pm$0.5} & \textbf{24.7$\pm$1.5} & \textbf{43.3$\pm$0.9} & \textbf{24.8$\pm$0.4} & {60.1$\pm$0.2} & \textbf{86.3$\pm$0.5} & \textbf{57.5$\pm$0.7} & \textbf{70.9$\pm$0.4} & \textbf{78.3$\pm$0.8} & \textbf{70.2$\pm$0.5}
 \\ \bottomrule[1.5pt]
\end{tabular}
}
\caption{Node classification accuracy (Macro F1-score).}
\label{tab:node_cls}
\end{table*}
\subsection{STAARW}\label{SpatioTemporal Principal Pathway Random Walk}
In this section, we extend Random Walk with Restart~\cite{Nassar2015StrongLI} to SpatioTemporal Activity-Aware Random Walk (STAARW),  to generate spatially and temporally node-to-node scores. Our approach draws inspiration from~\cite{Lee2022TimeawareRW} but differs in that STAARW selectively adopts a walking strategy biased by the node activity coefficient.

As depicted in Figure~\ref{STAA}(b), STAARW connects the same nodes from $\mathcal{G}_t$ to $\mathcal{G}_{t+1}$ at each time step $t$. Nodes not only move on the current snapshot $\mathcal{G}_t$ but also can leap to the same position in the next snapshot $\mathcal{G}_{t+1}$, allowing the random walk to gain temporal awareness. STAARW uses the node activity coefficients obtained from Node Evaluation to guide the node selection walking strategy. Initially, the wanderer starts from a seed node $s$ at the initial time step (e.g., $t = 1$). After a few moves, suppose the wanderer is at node $u$ in $\mathcal{G}_t$, it will take one of the following actions: \textbf{Action 1) Random walk:} Randomly moves to one of the neighbors of node $u$ in the current graph $\mathcal{G}_t$ with probability $1-\alpha-\beta_{t,u}$. \textbf{Action 2) Restart:} Goes back to the seed node $s$ in $\mathcal{G}_t$ with probability $\alpha$. \textbf{Action 3) Time travel:} Does time travel from node $u$ in $\mathcal{G}_t$ to node $u$ in $\mathcal{G}_{t+1}$ with probability $\beta_{t,u}$.

Here, $\alpha$ is the restart probability, and $\beta_{t,u}$ is the spatiotemporal activity coefficient of node $u$ at time $t$, with an upper limit of $1-\alpha$. Note that nodes cannot move backward from $\mathcal{G}_{t+1}$ because the present cannot influence the past.

Through STAARW, obtaining the stationary probability vector $\mathbf{x}_t \in \mathbb{R}^n$ of the wanderer visiting each node starting from the seed node $s$ on $\mathcal{G}_t$ recursively can be represented as:
\begin{equation}\label{eq7}
    \mathbf{x}_{t,s} = {\tilde{\mathbf{A}_t}}^{\top}(\mathcal{I}_n-\mathbf{\alpha}\mathcal{I}_n-\mathbf{\beta}_{t,\wedge})\mathbf{x}_{t,s} + \mathbf{\alpha}\mathbf{i}_s+\mathbf{\beta}_{t,\wedge}\mathbf{x}_{t-1,s},
\end{equation}
where $\beta_{t,\wedge}$ denote the diagonal matrices with main diagonal entries $ (\beta_{t,1},\beta_{t,2},\dots,\beta_{t,n})$. $\mathbf{i}_s$ is the $s$-th unit vector of size $n$. $\tilde{\mathbf{A}_t}$ is a row-normalized matrix of $\mathbf{A}_t$ (i.e., $\tilde{\mathbf{A}_t}=\mathbf{D}_t^{-1} \mathbf{A}_t$, where $\mathbf{A}_t$ is a self-looped adjacency matrix and $\mathbf{D}_t$ is a diagonal out-degree matrix of $\mathbf{A}_t$). We define $\mathbf{x}_{0, s}$ as $\mathbf{i}_s$. $\mathbf{\beta}_{t,\wedge}\mathbf{x}_{t-1,s}$ is the time travel term. The higher the activity coefficient $\beta_{t,u}$, the greater the time travel probability of node $u$, indicating a focus on earlier edges rather than recent edges.

In Equation (\ref{eq7}), $\mathrm{x}_{t, s} \in \mathbb{R}^{n \times 1}$ is a column vector of a probability distribution with respect to a seed node $s$. For all seeds $s \in \mathcal{V}$, $\{\mathrm{x}_{t, s}\}$ are horizontally stacked to form $\mathcal{X}_t \in \mathbb{R}^{n \times n}$, such that $\mathbf{x}_{t, s}$ is the $s$-th column of $\mathcal{X}_t$, i.e., $\mathrm{x}_{t, s}=\mathcal{X}_t \mathbf{i}_s$. According to Equation (\ref{eq7}), the following expression can be derived:
\begin{equation}\label{eq8}
\begin{aligned}
\mathbf{x}_{t,s} &= \mathbf{L}^{-1}_{t}({\alpha}\mathbf{i}_s+\beta_{t,\wedge}\mathbf{x}_{t-1,s}) \\
&= ( \mathbf{\alpha}\mathbf{L}^{-1}_{t} +  \mathbf{L}^{-1}_{t}\beta_{t,\wedge}\mathcal{X}_{t-1})\mathbf{i}_s = \mathcal{X}_t\mathbf{i}_s,
\end{aligned}
\end{equation}
where $\mathbf{L}_t = \mathcal{I}_n - {\tilde{\mathbf{A}_t}}^{\top}(\mathcal{I}_n-\mathbf{\alpha}\mathcal{I}_n-{\beta}_{t,\wedge})$, $\mathbf{x}_{t-1,s} = \mathcal{X}_{t-1} \mathbf{i}_s$, $\mathcal{X}_t=\mathbf{\alpha}{L}_t^{-1} +  \mathbf{L}^{-1}_{t}\beta_{t,\wedge}\mathcal{X}_{t-1}$ for $t>0$, and $\mathcal{X}_0=\mathcal{I}_n$. We refer to $\mathcal{X}_t$ as a SpatioTemporal Activity-Aware Random Walk Diffusion matrix at time $t$. A filtering threshold is used to set the values of $\mathcal{X}_t$ below $\rho$ to zero, thereby sparsifying $\mathcal{X}_t$.

\section{Experiment}
\subsection{Experimental Setup}
\textbf{Dataset.} 
We conduct experiments on seven public datasets extensively evaluated in dynamic graph representation learning studies. For link prediction tasks, we use the Bitcoin transaction network BitcoinAlpha~\cite{Kumar2016EdgeWP}, the Wikipedia admin voting network WikiElec~\cite{Leskovec2010PredictingPA}, and the hyperlink network between Reddit subforums RedditBody~\cite{Kumar2018CommunityIA}. For node classification tasks, we use the following datasets evaluated in~\cite{Xu2019SpatioTemporalAR}: the brain tissue connection network Brain, the co-author networks DBLP-3 and DBLP-5 from the DBLP database, and the post network Reddit. We ensure fair comparison and reproducibility by adopting the standard snapshot partitions from~\cite{lee2023time}.


\textbf{Baseline Methods.} 
We compare STAA with the following graph augmentation baselines: NONE (no augmentation), DropEdge (randomly removes edges at each epoch), GDC (employs Personalized PageRank), MERGE (combines adjacency matrices from time \textit{1} to $t$), TIARA~\cite{lee2023time}, and TGAC~\cite{Chen2023TemporalGR}. TIARA and TGAC are dynamic graph augmentation methods based on the assumption of temporal locality. Due to the lack of open-source implementations, we reproduce the graph augmentation method of TGAC (the version using degree centrality). For DropEdge and GDC, specific implementation details are unavailable, so we use experimental results from~\cite{lee2023time} to ensure a fair comparison.

We use GCN~\cite{Kipf2016SemiSupervisedCW} and two dynamic GNNs to perform dynamic graph tasks: GCRN~\cite{Seo2016StructuredSM} and EvolveGCN~\cite{Pareja2019EvolveGCNEG}. We apply static GCN to each graph snapshot to verify the informational value of temporal information. GCRN and EvolveGCN are classic and widely applied and studied dynamic GNNs~\cite{you2022roland,zhu2022high,mi2024dergcn}, and we utilize the implementations provided by ~\cite{Rozemberczki2021PyTorchGT}.

\textbf{Implementation Details.} 
For STAA, $\lambda$ is fixed at 1, $r$ at 6, $\epsilon$ at $10^{-8}$, and $\rho$ is searched in $[0.0001,0.01]$. $\alpha$, $\delta$, $\gamma$, and $W$ are chosen from $(0,1)$, $(0,2]$, $(0,1)$, and $[1,10]$, respectively. Adam optimizer is used with weight decay $10^{-4}$, learning rate in $[0.01,0.05]$, decay factor 0.999, and dropout ratio in $[0,0.5]$. Experiments are conducted 5 times, reporting mean and standard deviation of test values. PyTorch and DGL~\cite{Wang2019DeepGL} are used to implement all methods. We use Intel Xeon Silver 4310 as CPU and NVIDIA GeForce RTX 4090 as GPU.

\subsection{Link Prediction}
This task predicts edge existence at time $t+1$ using information up to time $t$. Following~\cite{Pareja2019EvolveGCNEG}, we split the time snapshots according to 70\% for training, 10\% for validation, and 20\% for testing. Equal numbers of negative (non-edge) and positive (edge) instances are sampled each time, with AUC as the metric. The number of epochs is 200, with an early stopping criterion of 50 epochs.

Table~\ref{tab:link_pre} shows that STAA improves GNNs performance across all datasets compared to NONE (no augmentation), while static augmentations like DropEdge and GDC do not, suggesting that spatial enhancement alone is ineffective for this task. STAA also surpasses TIARA and TGAC, suggesting its spatiotemporal enhancement is more effective for dynamic graph learning. Compared to methods based on the assumption of temporal locality, this superior performance highlights the effectiveness of STAA's strategy in suppressing active nodes' tendency towards recent edges, thereby reducing the weight of noisy edges.
\subsection{Node Classification}
This task classifies node labels in a dynamic graph, predicting node categories in the final snapshot. Following~\cite{Xu2019SpatioTemporalAR}, nodes are split into training, validation, and test sets (7:1:2). Node embeddings go through a softmax classifier, and Macro F1-score is used due to label imbalance. The number of epochs is set to 1,000 with an early stopping patience of 100.

Table~\ref{tab:node_cls} shows that STAA improves GNNs performance across all datasets, particularly on Brain. In contrast, TIARA and TGAC, which are based on the temporal locality assumption, are inferior to STAA. Notably, TGAC achieves lower accuracy than None (without augmentation) on nearly all datasets, suggesting that TGAC likely amplifies noise rather than enhancing effective spatiotemporal information.

\section{Acknowledgment} This work is supported by the National Key R\&D Program of China [2022YFF0902703].

\section{Conclusion}
To improve dynamic graph learning, we reveal the spatiotemporal activity of nodes using graph wavelet transform and propose a dynamic graph augmentation method called STAA. STAA guides random walk strategy selection using node activity coefficients, reducing noisy edge weights and enhancing spatiotemporal information. Through extensive experimental analysis on multiple public datasets, we demonstrate that STAA performs better in enhancing GNNs for link prediction and node classification tasks on dynamic graphs.

\newpage
\bibliographystyle{IEEEtran.bst}
\bibliography{ref.bib}

\end{document}